\documentclass[sigconf]{acmart}
\usepackage{amsfonts}
\usepackage{amsmath,amsfonts}
\usepackage{amssymb}
\usepackage{makecell}
\usepackage{mathrsfs}
\usepackage{graphicx}
\usepackage{algorithmic}
\usepackage{multirow}
\usepackage{booktabs}
\usepackage{xcolor}
\AtBeginDocument{%
  }

\setcopyright{acmlicensed}
\copyrightyear{2024}
\acmYear{2024}

\acmConference[MM '24] {Proceedings of the 32nd ACM International Conference on Multimedia}{October 28--November 1, 2024}{Melbourne, VIC, Australia.}
\acmBooktitle{Proceedings of the 32nd ACM International Conference on Multimedia (MM '24), October 28--November 1, 2024, Melbourne, VIC, Australia}
\acmDOI{10.1145/3664647.3680718}
\acmISBN{979-8-4007-0686-8/24/10}

\begin{document}

\title{MoTrans: Customized Motion Transfer with Text-driven Video Diffusion Models}

\author{Xiaomin Li}
\orcid{https://orcid.org/0000-0001-7202-6865}
\affiliation{%
  \institution{Dalian University of Technology}
  \city{Dalian}
  \country{China}}
\email{xmli22@mail.dlut.edu.cn}

\author{Xu Jia}
\authornote{Corresponding author.}
\orcid{https://orcid.org/0000-0003-3168-3505}
\affiliation{%
  \institution{Dalian University of Technology}
  \city{Dalian}
  \country{China}}
\email{jiayushenyang@gmail.com}

\author{Qinghe Wang}
\orcid{https://orcid.org/0000-0001-6908-5485}
\affiliation{%
  \institution{Dalian University of Technology}
  \city{Dalian}
  \country{China}}
\email{qinghewang@mail.dlut.edu.cn}

\author{Haiwen Diao}
\orcid{https://orcid.org/0000-0002-4156-5417}
\affiliation{%
  \institution{Dalian University of Technology}
  \city{Dalian}
  \country{China}}
\email{diaohw@mail.dlut.edu.cn}

\author{Mengmeng Ge}
\orcid{https://orcid.org/0009-0003-1301-1323}
\affiliation{%
  \institution{Tsinghua University}
  \city{Beijing}
  \country{China}}
\email{gmm21@mails.tsinghua.edu.cn}

\author{Pengxiang Li}
\orcid{https://orcid.org/0009-0005-6906-7005}
\affiliation{%
  \institution{Dalian University of Technology}
  \city{Dalian}
  \country{China}}
\email{lipengxiang@mail.dlut.edu.cn}

\author{You He}
\orcid{https://orcid.org/0000-0002-6111-340X}
\affiliation{%
  \institution{Tsinghua University}
  \city{Beijing}
  \country{China}}
\email{heyou_f@126.com}

\author{Huchuan Lu}
\orcid{https://orcid.org/0000-0002-6668-9758}
\affiliation{%
  \institution{Dalian University of Technology}
  \city{Dalian}
  \country{China}}
\email{lhchuan@dlut.edu.cn}

\renewcommand{\shortauthors}{Xiaomin Li et al.}

\begin{abstract}
Existing pretrained text-to-video~(T2V) models have demonstrated impressive abilities in generating realistic videos with basic motion or camera movement. However, these models exhibit significant limitations when generating intricate, human-centric motions. Current efforts primarily focus on fine-tuning models on a small set of videos containing a specific motion. They often fail to effectively decouple motion and the appearance in the limited reference videos, thereby weakening the modeling capability of motion patterns.
To this end, we propose MoTrans, a customized motion transfer method enabling video generation of similar motion in new context. Specifically, we introduce a multimodal large language model~(MLLM)-based recaptioner to expand the initial prompt to focus more on appearance and an appearance injection module to adapt appearance prior from video frames to the motion modeling process.
These complementary multimodal representations from recaptioned prompt and video frames promote the modeling of appearance and facilitate the decoupling of appearance and motion. 
In addition, we devise a motion-specific embedding for further enhancing the modeling of the specific motion.
Experimental results demonstrate that our method effectively learns specific motion pattern from singular or multiple reference videos, performing favorably against existing methods in customized video generation.
\end{abstract}


\begin{teaserfigure}
  \centering
  \includegraphics[width=0.99\textwidth]{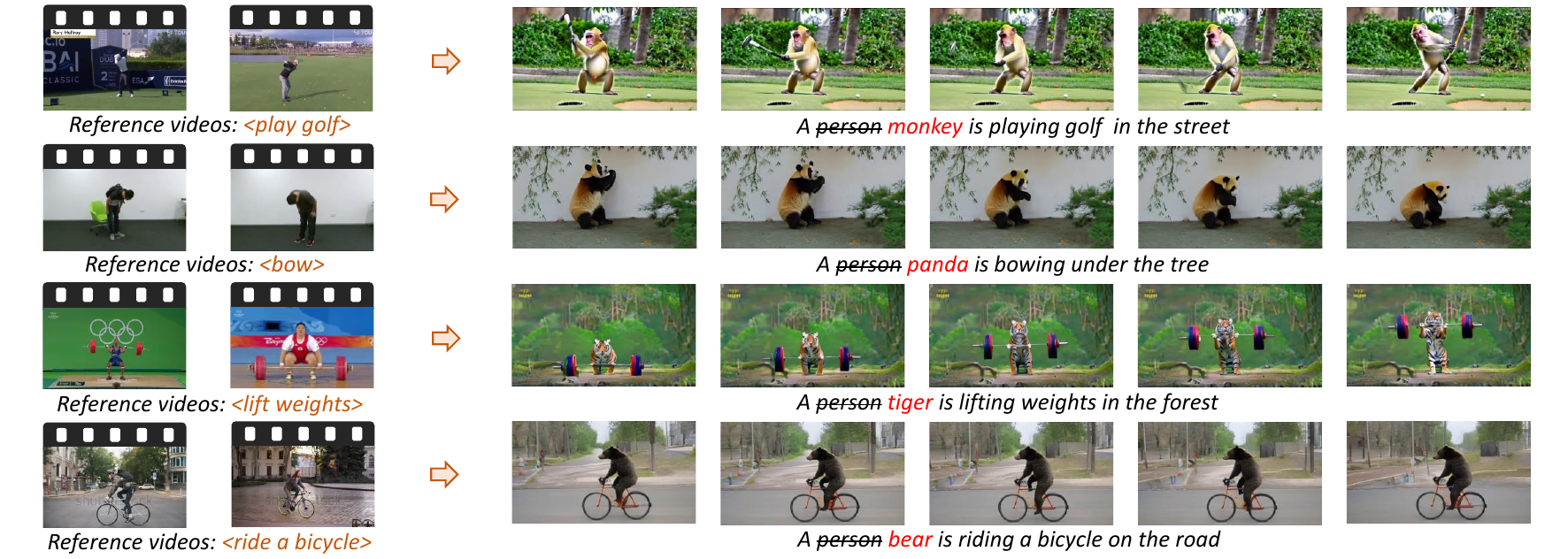}
  \caption{MoTrans is meticulously crafted to capture precise motion patterns from either singular or multiple reference videos, facilitating seamless transfer of these motions onto fresh subjects within diverse contextual scenes.}
  \label{fig:teaser}
\end{teaserfigure}


\maketitle

\section{Introduction}
Diffusion-based video generation has achieved significant breakthroughs~\cite{pika,walt,gen2,sora}, facilitating the production of high-quality, imaginative videos. 
While foundation Text-to-Video~(T2V) models can generate diverse videos from provided text, tailoring them to generate specific motion could more closely align with user's preferences. Akin to subjects customization in Text-to-Image~(T2I) tasks~\cite{dreambooth,anydoor,cones2}, human-centric motions in videos can also be customized and transferred to various subjects, which holds significant practical benefits for animation and film production~\cite{tune-a-video,dreamvideo}.

Existing pretrained T2V models~\cite{modelscope,zeroscope} often struggle to generate intricate, human-centric motions like golf swings and skateboarding, which involve multiple continuous sub-motions. 
One potential reason is that these foundation models are predominantly trained on highly diverse datasets~\cite{webvid} sourced from the internet, which may have imbalanced data distribution.
As a result, the models might encounter certain motions infrequently, leading to inadequate training for those motions.
To better generate particular motions, these pretrained T2V models~\cite{modelscope,zeroscope} require fine-tuning on a small set of videos containing the desired motion pattern. 
However, fine-tuning the model directly without any additional constraints is prone to lead to an undesirable coupling between the motion and the appearance in the limited reference videos and weakening the modeling capability of motion patterns.


Several works~\cite{motiondirector,dreamvideo,customize-a-video,motioncrafter} have been proposed to address the issue outlined above. These approaches predominantly leverage a dual-branch architecture, with one branch dedicated to capturing single-frame spatial information and the other to inter-frame temporal dynamics. Additionally, they also introduce decoupling mechanisms, such as embedding appearance priors to guide the focus of temporal layers on motion~\cite{dreamvideo} or adjusting latent codes to minimize the negative impact of appearance~\cite{motiondirector}.
Despite their efforts to separate appearance from motion, these approaches exhibit insufficient learning of motion patterns, resulting in videos with diminished motion magnitudes and a deviation from the motion observed in reference videos to some extent.

To this end, we introduce \textbf{MoTrans}, a customized \textbf{Mo}tion \textbf{Trans}-fer method, which mainly focuses on modeling the motion patterns in reference videos while avoiding overfitting to its appearance. Specifically, we adopt a two-stage training strategy, with an appearance learning stage and a motion learning stage respectively modeling appearance and motion. To alleviate the coupling issue between appearance and motion, we undertake comprehensive explorations in both stages.
1) During the appearance learning stage, a multimodal large language model~(MLLM) is adopted as the recaptioner to expand the original textual descriptions of the reference video. 
2) During the motion learning stage, before adapting the temporal module to a specific motion, representations of video frame are pre-injected to compel this module to capture motion dynamics.
The complementary multimodal information from expanded prompt and video frame promotes the modeling of appearance and decomposition of appearance and motion.  
Notably, it has been observed that motions in videos are primarily driven by verbs within the prompt. Inspired by this observation, we employ a residual embedding to enhance the token embeddings of the verbs corresponding to motion, thereby capturing the specified motion patterns in the reference video. 
Extensive experimental results demonstrate that our method effectively mitigates the issue of overfitting to appearance and produces high-quality motion, performing favorably against other state-of-the-art methods.
The main contributions of this paper can be summarized as follows:
\begin{itemize}
    \item We propose MoTrans, a customized video generation method enabling motion pattern transfer from single or multiple reference videos to various subjects.
    \item By introducing an MLLM-based recaptioner and appearance prior injector, we leverage complementary text and image multimodal information to model the appearance information, effectively mitigating the issue of coupling between motion and the limited appearance.
    \item We introduce the motion-specific embedding, which is integrated with temporal modules to collaboratively represent specific motion within reference videos.
    \item Experimental results demonstrate that our method surpasses other motion customization methods, enabling any motion customization contextualized in different scenes.
\end{itemize}
\section{Related Work}
\begin{figure*}[!t]
\centering
\includegraphics[width=1\textwidth]{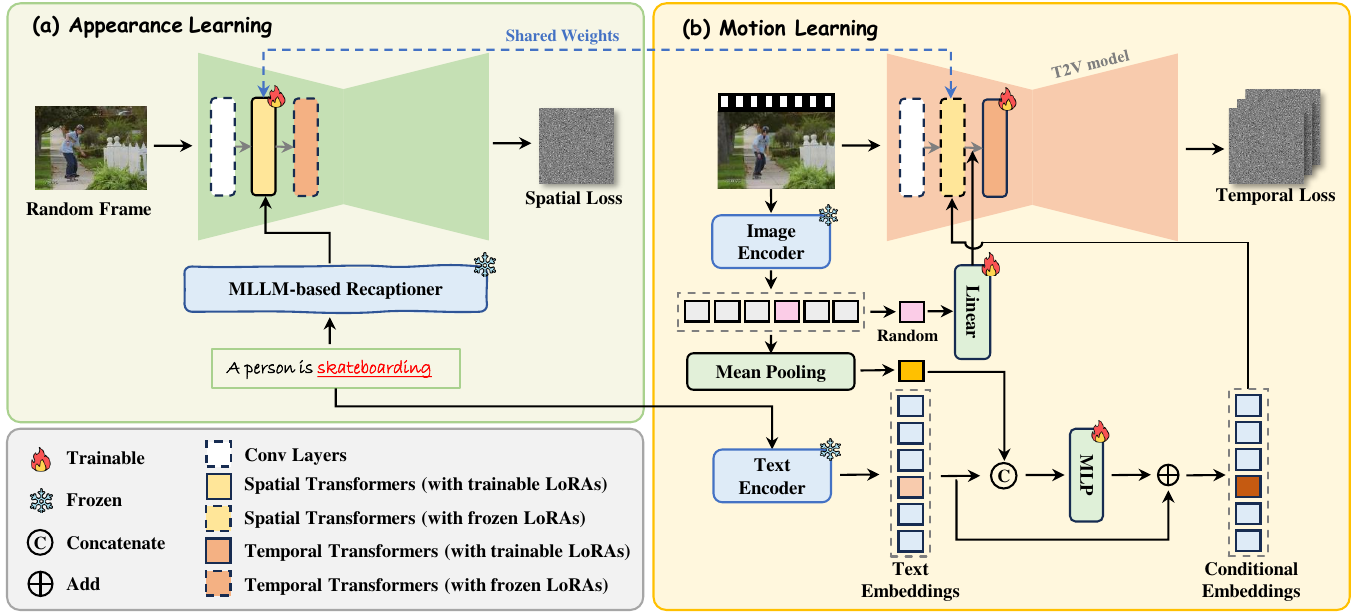}
\caption{Overview of the proposed MoTrans. In the appearance learning stage, an MLLM-based recaptioner is employed to extend the base prompt, encouraging the spatial LoRAs to sufficiently learn appearance information. The weights of spatial LoRAs are shared in the second stage. In the motion learning stage, video frame embeddings are injected as appearance priors, compelling the temporal LoRAs to concentrate on motion learning. Furthermore, we adopt MLP to learn a motion-specific embedding, which is jointly trained with the temporal LoRAs to fit specific motion patterns in the reference video.}
\label{fig:overall_framework}
\end{figure*}

\subsection{Text-to-Video Generation}
Diffusion models are catalyzing rapid advancements in image generation tasks~\cite{ldm,dit,pixartalpha} and have spawned numerous valuable applications~\cite{textualinversion,stableidentity,nulltext,guided,unipaint,paint-by-example}.
This success has garnered significant interest in extending these capabilities to video generation~\cite{lavie,animatediff,tune-a-video,conditionvideo}.
Early efforts in T2V domain~\cite{imagen_video,vdm,make-a-video,lavie} primarily focus on cascading video frame interpolation and super-resolution models to generate high-resolution videos, which seems to be complex and cumbersome. In contrast, ModelScopeT2V~\cite{modelscope} represents a significant shift by incorporating spatio-temporal blocks atop stable diffusion ~\cite{ldm} to model motion more effectively.     
Building on this, ZeroScope~\cite{zeroscope} expands the training data and utilizes watermark-free data for fine-tuning, enabling the generation of videos with improved resolution and enhanced quality. 

Recently, a new wave of high-quality T2V models~\cite{emuvideo,walt,pixeldance,magicvideo} has achieved impressive progress.
VideoCrafter2~\cite{videocrafter2} utilizes low-quality videos to ensure motion consistency while employing high-quality images to enhance video quality and conceptual composition ability.
Commercial models such as Pika~\cite{pika} and Gen-2~\cite{gen2} also exhibit exceptionally strong generative capabilities. 
Moreover, OpenAI's recent launch of the Sora model~\cite{sora}, capable of generating high-quality videos up to 60 seconds in length, marks a significant milestone in video generation.
Although the above foundation T2V models can generate appealing videos, they face challenges in precisely controlling the generated motion.

\subsection{Customized Video Generation}
Existing T2V models~\cite{videocrafter2, zeroscope, modelscope} excel at generating simple motions or camera movements, struggling to produce specific human-centric motions that align with user preferences. To this end, some models have been introduced to synthesize specific motion pattern and transfer it to diverse subjects.
For customized motion transfer~\cite{animate_anyone,magicanimate,magicdance}, some methods employ additional pose maps~\cite{openpose} or dense poses~\cite{densepose} as guidance and require substantial amounts of training data. 
These approaches allow the production of animations without any need for fine-tuning once they are adequately trained.
However, they primarily focus on human-to-human motion transfer and often struggle to transfer motion to subjects that significantly deviate from the human domain, such as animals. 

We aim to learn specific motion patterns rather than precisely replicate every frame's action. This task~\cite{motiondirector,dreamvideo,lamp,motioncrafter,customizing_motion,customize-a-video} requires only a minimal amount of training data sharing the same motion concept. 
Similar to the T2I method DreamBooth~\cite{dreambooth}, these approaches necessitate individual training for each type of motion. Since the generation process does not require additional control conditions such as pose, the resulting motions are more flexible and do not need to follow each frame of the reference video strictly. MotionDirector
~\cite{motiondirector} learns both camera movement and motion, adopting a dual-path way framework to separately learn appearance and motion. During motion learning, the spatial layers trained for appearance learning are frozen to inhibit the temporal layers from learning appearance. 
DreamVideo~\cite{dreamvideo} introduces structurally simple identity and motion adapters to learn appearance and motion, respectively.
To decouple spatial and temporal information, it proposes injecting appearance information into the motion adapter, forcing the temporal layers to learn motion.

Although some methods~\cite{motiondirector, motioncrafter,dreamvideo,customize-a-video} realize the issue of appearance-motion coupling, they are still prone to synthesizing videos overfitting to the appearances of training data to a certain extent, thereby exhibiting insufficient learning of motion patterns. Furthermore, some methods~\cite{lamp,motioncrafter} learn motions that are easier to model. In this paper, we are more concerned with challenging actions with larger ranges of motion, such as sports actions.

\begin{figure}[t]
\begin{center}
\includegraphics[width=0.92\linewidth]{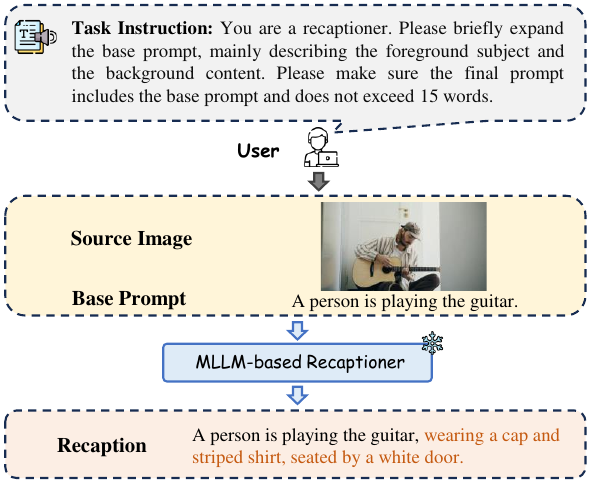}
\end{center}
   \caption{Illustration of multimodal recaptioning. Given an image, an MLLM-based recaptioner is employed to expand the base prompt according to the task
instruction, enabling the extended prompt to fully describe its appearance.} 
\label{fig:prompt_expander}
\end{figure}

\section{Method}
\subsection{Overview}
\begin{figure}[t]
\begin{center}
\includegraphics[width=0.88\linewidth]{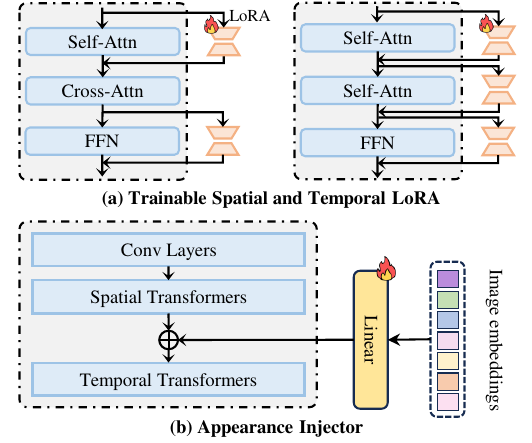}
\end{center}
   \caption{Details of trainable LoRAs and appearance injector. (a) Parameters of the base model are frozen and only parameters of LoRAs are updated. (b) The image embedding is processed through a Linear layer before being fused with the hidden states from the spatial transformers. This pre-injected appearance prior encourages the temporal LoRAs to capture motion patterns effectively.}
\label{fig:injector}
\end{figure}

Given a single video or multiple videos with similar motions, the goal of our task is to learn the specific motion or the common motion pattern contained in reference videos. Subsequently, the learned motion can be adapted to new subjects contextualized in different scenes. 
As illustrated in Fig.~\ref{fig:overall_framework}, the overall training pipeline is divided into the appearance learning stage and the motion learning stage. 
In the appearance learning stage, we employ an MLLM-based recaptioner to expand the initial prompt of the reference videos. It could promote the modeling capabilities of the spatial attention modules for appearance information.
At this stage, we only train spatial low-rank adaptions (LoRAs) and share the weights with the second stage to fit the appearance of the corresponding reference video.
As shown in Fig.~\ref{fig:injector}~{(a)}, to preserve the textual alignment capability of the pretrained T2V model, we freeze the parameters of the cross-attention layer and only inject LoRAs into the self-attention and feed-forward layers~(FFN). In the motion learning stage, before adapting temporal modules to a specific motion, image embeddings are injected to introduce appearance priors, thereby forcing the temporal LoRAs to focus on motion modeling. Additionally, we employ a multilayer perceptron~(MLP) to augment the token embeddings corresponding to verbs, which is jointly trained with temporal LoRAs to capture specific motion pattern.
For temporal modules, LoRAs are injected into both the self-attention layer and FFN of the temporal transformers.

During the inference stage, we integrate the temporal LoRAs and the residual embedding into pretrained video generation models to transfer the specific motion to new subjects.

\subsection{Multimodal Appearance-Motion Decoupling}
The primary objective of our task is to learn motion patterns specified by several reference videos. Due to the inherent characteristics of the diffusion model's training loss, the leakage of some appearance information is inevitable in the motion learning stage. To separate motion from appearance to a certain extent during the motion learning process, we propose an MLLM-based recaptioner and an appearance injector. In this manner, the complementary multimodal appearance information provided by text and video facilitates the decoupling of appearance and motion information. In general, complementary multimodal information is often beneficial for downstream tasks~\cite{vipt}.

\noindent
\textbf{MLLM-based recaptioner.} 
MLLMs like LLaVA 1.5~\cite{llava} or GPT4~\cite{chatgpt} have robust in-context reasoning and language understanding capabilities, which can be used for image recaptioning~\cite{dalle3,rpg}. 
As illustrated in Fig.~\ref{fig:prompt_expander}, let $\mathcal{V}=\{f^i|i=1,...,l\}$ denote the reference video with $l$ frames,
given a carefully crafted task instruction, the recaptioner can perform text-to-text translation and expand the base prompt $\mathbf{c}_b$ based on a random frame $f^i$. In this manner, the recaptioned prompt $\mathbf{c}_r$ can comprehensively describe the appearance information contained within the video frames.
Through training, the spatial attention module will adapt to the appearance information of the reference video and remains frozen in the subsequent stage, encouraging the temporal attention module to effectively model the motion information in the videos.


During the appearance learning stage, we adopt a frozen MLLM and only spatial LoRAs need to be trained. The optimization process of this stage is defined as follows:
\begin{equation}
    \label{eq:app_loss}
    \mathcal{L}_{s} = \mathbb{E}_{\mathbf{z_0^i},\mathbf{c}_r,\mathbf{\epsilon} \sim \mathcal{N}(0,\mathit{I}),t}[||\mathbf{\epsilon}-\mathbf{\epsilon}_\theta(\mathbf{z_t^i}, \tau_\theta(\mathbf{c}_r), t)||^2_2].
\end{equation}
Here, a VQ-VAE~\cite{vqvae} initially compresses frame $f^i$ into a latent representation $\mathbf{z_0^i} \in \mathbb{R}^{b \times 1 \times h \times w \times c}$, where $b,h,w,c$ represent batch size, height, width, and channel count, respectively. $\mathbf{z_t^i}$ is the noised latent code at timestep $t \sim \mathcal{U}(0,T)$. $\tau_\theta(\cdot)$ denotes the pretrained OpenCLIP ViT-H/14~\cite{clip} text encoder. Meanwhile, the network $\mathbf{\epsilon}_\theta(\cdot)$ is trained to predict the noises added at each timestep.

\noindent
\textbf{Appearance injector.}
In addition to leveraging recaptioned prompts to enhance the modeling of appearance, integrating embedding information from video frames themselves can also yield significant benefits. These two modalities collaboratively contribute to the effective decomposition of motion from its appearance.
Drawing inspiration from~\cite{dreamvideo}, we inject appearance information in the second stage to diminish its impact on motion learning.
As shown in Fig.~\ref{fig:overall_framework}~(b), an image encoder $\mathbf{\psi}$ is utilized to obtain embeddings of all video frames, we randomly select an image embedding $\mathbf{\psi (f^i)} \in \mathbb{R}^{1 \times d}$ from the input video, where $d$ denotes the dimension of image embedding. Then the appearance information is injected before the temporal transformers, as demonstrated in Fig.~\ref{fig:injector}~(b).
Formally, for each UNet block $l$, the spatial transformer produces the hidden states $h_s^l \in \mathbb{R}^{(b \times h \times w)\times f \times c}$. We employ a linear projection to broadcast the input embeddings across all frames and spatial positions, which are then summed with the hidden states $h_s^l$ before being fed into the temporal transformer. In this way, the appearance representations from the visual modal are pre-injected. The entire process can be formulated as follows: 
\begin{equation}
    \label{temporal_hidden_states}
    h_t^l=h_s^l \odot (W_{p}\cdot \mathbf{\psi(f^i)}),
\end{equation}
where $W_{p}$ represents the weights of the linear projection layer, with its output dimension adapting to variations in the dimensions of the UNet hidden states. And $\odot$ denotes the broadcast operator.

\subsection{Motion Enhancement}
An intuitive observation is that motion patterns in videos generally align with verbs in a text prompt. Hence, we posit that emphasizing verbs could potentially encourage the model to enhance its modeling of motion in the reference videos.

\noindent
\textbf{Motion Enhancer.} 
A heuristic strategy for enhancing the modeling of motion involves leveraging visual appearance information to enrich the textual embedding representation of motion concept. This is achieved by learning a residual motion-specific embedding on top of the base text embedding. The base embedding can be considered a coarse embedding corresponding to a general motion category, whereas the residual embedding is tailored to capture the specific motion within given reference videos.

Specifically, we employ a pretrained text encoder~$\tau_\theta$ to extract text embeddings from a sequence of words~$\mathcal{S}=\{s_1,...,s_N\}$. To locate the position $\mathit{i}$ of the verb $s_i$ in the text prompt, we use Spacy for part-of-speech tagging and syntactic dependency analysis. 
Following this, the base motion embeddings corresponding to the motion concept are then selected. 
As shown in Fig.~\ref{fig:overall_framework}~(b), 
video frames are initially processed by an image encoder to generate frame-wise embeddings. To capture temporal interactions, these embeddings are aggregated through a mean pooling operation, resulting in a unified video embedding. 
This video embedding is concatenated with the base motion embedding and further processed by an MLP. The MLP comprises two linear layers separated by a Gaussian Error Linear Unit~(GELU)~\cite{gelu} activation function.
Subsequently, we compute a residual embedding, which is added to the base motion embedding to form an enhanced motion-specific representation. 
Mathematically, let $E_b$ and $E_r$ represent the base embedding and learnable residual embedding, respectively, the operation can be expressed as follows:
\begin{equation}
    \label{eq:mlp}
    E_r=W_2\cdot(\sigma_{GELU}(W_1\cdot([MeanPool(\mathbf{\psi(\mathcal{V})}),\mathbf{\tau_\theta(s_i)}]))),
\end{equation}
\begin{equation}
    \label{eq:condition}
    E_{cond}=E_b+E_r.
\end{equation}
Here, $[\cdot]$ refers to the concatenation operation, and $W_1$ and $W_2$ denote the weights of two Linear layers in MLP. The GELU function is represented by $\sigma_{GELU}$. This motion-specific embedding is integrated with the text embeddings of other words in the prompt to serve as the new condition for training temporal transformers. 

To prevent the learned residuals from becoming excessively large, akin to the strategy in~\cite{e4t}, we introduce an L2 regularization term as a constraint as:
\begin{equation}
    \label{eq:loss_reg}
    \mathcal{L}_{reg}=||E_{r}||^2_2
\end{equation}
Similar to the appearance learning stage, the loss function in the motion learning stage calculates the Mean Squared Error (MSE) loss between the predicted noise of the diffusion model and the ground truth noise, except that the frame dimension is no longer 1. Therefore, the final loss function for this stage is defined as:
\begin{equation}
    \label{eq:loss_temp}
    \mathcal{L}_t = \mathbb{E}_{\mathbf{z_0^{1:N}},\mathbf{c}_b,\mathbf{\epsilon} \sim \mathcal{N}(0,\mathit{I}),t}[||\mathbf{\epsilon}-\mathbf{\epsilon}_\theta(\mathbf{z_t^{1:N}}, \tau_\theta(\mathbf{c}_b), t)||^2_2].
\end{equation}
For motion learning, the loss function is the combination of temporal loss and a constraint term as follows,
\begin{equation}
    \label{eq:loss_motion}
    \mathcal{L}_{motion}=\mathcal{L}_{t}+\lambda\mathcal{L}_{reg},
\end{equation}
where $\lambda$ controls the relative weight of the regularization term.

\section{Experiments}
\begin{figure*}[!t]
\centering
\includegraphics[width=1\textwidth]{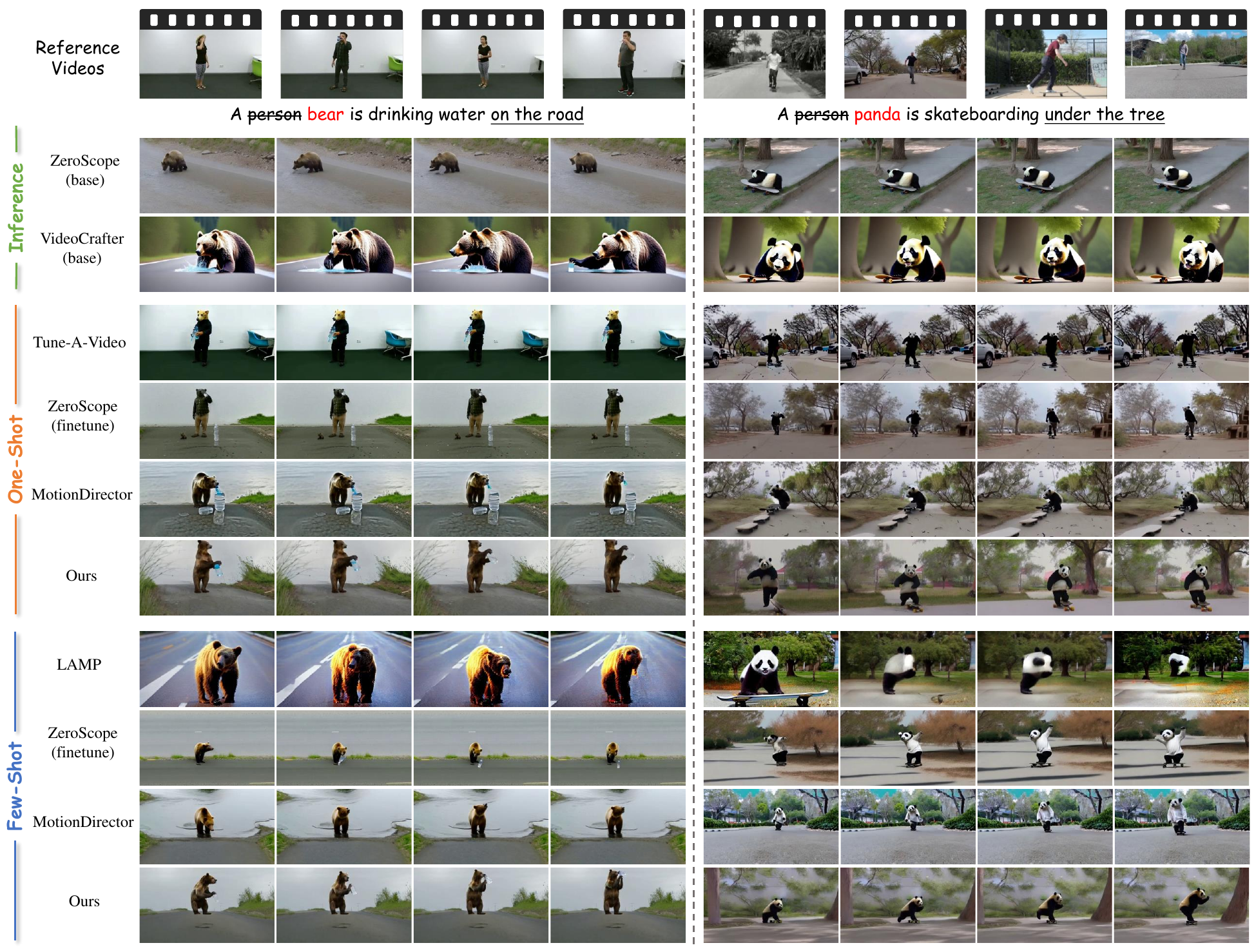}
\caption{Qualitative comparison of customized motion transfer. The reference videos on the left demonstrate the motion of a person slowly lifting their hand to drink water. On the right, the videos show a skateboarding pushing action, where the person pushes off the ground with their foot and then slides forward. For one-shot motion customization, the learned motion refers to the second example from the reference videos. \textit{Best viewed zoomed-in.}}
\label{fig:qualitative}
\end{figure*}

\subsection{Experimental Setup}
\noindent
\textbf{Dataset.} 
We collect a dataset that includes 12 distinct motion patterns, sourced from the Internet, the UCF101 dataset~\cite{ucf101}, the UCF Sports Action dataset~\cite{ucfsports}, and NTU RGB+D 120~\cite{ntuRGB+D120}. Each motion pattern is represented by approximately 4-10 training videos.
The dataset consists of various sports motions, such as weightlifting and golf swing,  alongside large-scale limb movements like waving hands and drinking water. 
For evaluation, we employ six base prompt templates that involve variations in subject, motion, and context. An example template is "A \{cat\} is \{motion\} \{in the living room\}", with placeholders indicating dynamic elements. Videos corresponding to each motion are generated based on these six prompt categories.
More details of our dataset are available in the supplementary materials.

\noindent
\textbf{Implementation details.} 
We employ ZeroScope as the base T2V model, which is trained with the AdamW~\cite{adamw} optimizer across approximately 600 steps with a learning rate of $5e-4$. For the spatial and temporal transformers, we specifically fine-tune LoRAs instead of all parameters, with the LoRA rank set to 32. 
The image encoder used for appearance injection is derived from OpenCLIP ViT-H/14, which is also used to calculate CLIP-based metrics.
The regularization loss coefficient for normalizing the verb's residual embedding is $1e-4$. 
During inference, we employ DDIM~\cite{ddim} sampler with 30-step sampling and classifier-free guidance scale~\cite{cfg} of 12. It takes about 19 seconds to generate 24-frame videos at 8 fps with a resolution of $576 \times 320$.
All experiments are conducted on a single NVIDIA A100 GPU.

\noindent
\textbf{Comparison methods.}
To investigate the generative capabilities of existing T2V models, we compare our approach with prominent open-source models, including ZeroScope~\cite{zeroscope} and VideoCrafter2~\cite{videocrafter2}. 
Additionally, we explore the effectiveness of directly fine-tuning ZeroScope on a small set 
of videos containing a specific motion. It is noteworthy that fine-tuning is not applied to the entire diffusion model but specifically targets the LoRAs within the temporal transformers.
Our proposed method is adaptable for both single and multiple video customization scenarios. Consequently, we benchmark our approach against open-source methods specialized for one-shot customization, such as Tune-a-Video~\cite{tune-a-video}, and for few-shot customization, like LAMP~\cite{lamp}. Further comparisons are conducted with MotionDirector, which serves as our baseline.


\noindent
\textbf{Evaluation metrics.}
The performance of the comparison methods is evaluated by four metrics. CLIP Textual Alignment~\textbf{(CLIP-T)} is employed to assess the correspondence between the synthesized video and the provided prompt, while Temporal Consistency~\textbf{(TempCons)} measures frame consistency within videos. To assess appearance overfitting observed in some comparison methods, we introduce CLIP Entity Alignment~\textbf{(CLIP-E)} metric, which is similar to CLIP-T but focuses on prompts containing only entities, like "a panda". This metric evaluates if the synthesized video accurately generates the entity specified by the new prompt. To the best of our knowledge, there exists no metric measuring the congruence between motion patterns in the synthesized videos and those in the reference videos. Therefore, we propose Motion Fidelity~\textbf{(MoFid)}, which is based on the video understanding model VideoMAE~\cite{videomae}. 
Specifically, a video $v_m^i$ is randomly selected from the training videos, and a pretrained VideoMAE $f(\cdot)$ is used to obtain the embeddings for both the selected video $v_m^i$ and the synthesized video $\bar{v}_k$. Formally, motion fidelity is calculated as follows:
\begin{equation}
    \label{eq:Motion Fidelity}
    \mathcal{E}_m = \frac{1}{|\mathcal{M}||\bar{v}_m|}\sum_{m\in \mathcal{M}}\sum_{k=1}^{|\bar{v}_m|}cos(f(v_m^i),\bar{v}_k),
\end{equation}
where $\mathcal{M}$ denotes the set of motions, $|\bar{v}_m|$ is the number of videos with motion $m$ in the generated videos, and $cos(\cdot)$ refers to cosine similarity function.
Actually, a good motion transfer model should generate the subject implied
by the prompts and perform the motion specified in the reference
videos. Therefore, it is necessary to evaluate the model’s capabilities by considering the MoFid and CLIP-T/CLIP-E metrics together.
\begin{table}[t]
\renewcommand\arraystretch{1.1}
\centering
\caption{Quantitative evaluation of customized motion transfer methods. The best results under one-shot and few-shot settings are highlighted in blue and red, respectively.
}
\label{tab:quantitative}
\resizebox{0.48\textwidth}{!}{%
\fontsize{26pt}{30pt}\selectfont
\begin{tabular}{lccccc}
\toprule
&  & \textbf{CLIP-T}~($\uparrow$)  &  \textbf{CLIP-E}~($\uparrow$)  & \textbf{TempCons~}~($\uparrow$)    & \textbf{MoFid}~($\uparrow$)  \\ 
\midrule
\multirow{2}{*}{Inference} &  ZeroScope~\cite{zeroscope}       & 0.2017 & 0.2116 & 0.9785 & 0.4419 \\
                           &  VideoCrafter~\cite{videocrafter2}    & 0.2090 & 0.2228 & 0.9691 & 0.4497 \\ \hline 
\multirow{4}{*}{One-shot}  &  Tune-a-video~\cite{tune-a-video}     & 0.1911 & 0.2031 & 0.9401 & 0.5627 \\
                           &  ZeroScope~(fine-tune) & 0.2088 & 0.2092 & 0.9878 & \textcolor{blue}{\textbf{0.6011}} \\
                           &  MotionDirector~\cite{motiondirector}    & 0.2178  & 0.2130 & \textcolor{blue}{\textbf{0.9889}} & 0.5423   \\
                           &  MoTrans~{(ours)}    & \textcolor{blue}{\textbf{0.2192}}  & \textcolor{blue}{\textbf{0.2173}} & 0.9872 & 0.5679  \\ 
\midrule
\multirow{4}{*}{Few-shot}  &  LAMP~\cite{lamp}  & 0.1773 & 0.1934 & 0.9587 & 0.4522 \\       
                           &  ZeroScope~(fine-tune)  & 0.2191 & 0.2132 & 0.9789 & 0.5409 \\
                           &  MotionDirector     & 0.2079 & 0.2137 & 0.9801 & 0.5417 \\
                           &  MoTrans~{(ours)}    & \textcolor{red}{\textbf{0.2275}}  & \textcolor{red}{\textbf{0.2192}} & \textcolor{red}{\textbf{0.9895}} & \textcolor{red}{\textbf{0.5695}}   \\
\bottomrule
\end{tabular}
}
\end{table}

\begin{table}[t]
\renewcommand\arraystretch{1.1}
\centering
\caption{Quantitative results of the ablation study.
}
\label{tab:ablation}
\resizebox{0.48\textwidth}{!}{%
\fontsize{26pt}{30pt}\selectfont
\begin{tabular}{lccccc}
\toprule
&  & \textbf{CLIP-T}~($\uparrow$)  &  \textbf{CLIP-E}~($\uparrow$)  & \textbf{TempCons
}~($\uparrow$)    & \textbf{MoFid}~($\uparrow$)  \\ 
\midrule
\multirow{4}{*}{One-shot}  &  w/o MLLM-based recaptioner & 0.2138 & 0.2101 & 0.9865 & 0.6129 \\
                           &  w/o appearance injector & 0.2114 & 0.2034 & 0.9862 & \color{blue}{\textbf{0.6150}} \\
                           &  w/o motion enhancer & 0.2164 & 0.2135 & 0.9871 & 0.5643 \\
                           &  MoTrans & \color{blue}{\textbf{0.2192}} & \color{blue}{\textbf{0.2173}} & \color{blue}{\textbf{0.9872}} & 0.5679 \\
\midrule
\multirow{4}{*}{Few-shot}  &  w/o MLLM-based recaptioner & 0.2179 & 0.2138 & 0.9792 & 0.5997 \\       
                           &  w/o appearance injector & 0.2143 & 0.2132 & 0.9807 & \color{red}{\textbf{0.6030}} \\
                           &  w/o motion enhancer & 0.2211 & 0.2171 & 0.9801 & 0.5541 \\
                           &  MoTrans & \color{red}{\textbf{0.2275}} & \color{red}{\textbf{0.2192}} & \color{red}{\textbf{0.9895}} & 0.5695 \\
\bottomrule
\end{tabular}
}
\end{table}

\subsection{Main Results}
\noindent
\textbf{Qualitative Evaluation.}
To validate the motion customization capabilities of our method, we conduct a comparative analysis with several representative open-source methods tailored for one-shot and few-shot motion customization. 
As depicted in Fig.~\ref{fig:qualitative}, direct inference using pretrained T2V models ZeroScope and VideoCrafter fails to synthesize specific motion patterns due to the lack of fine-tuning on specified videos. Additionally, the synthesized videos exhibit notably small motion amplitudes, suggesting that pretrained T2V models struggle to generate complex, human-centric motion. In particular, these models face significant challenges in generating specific motions without targeted training on specific videos.
Furthermore, unconstrained fine-tuning of Zeroscope leads to an undesirable coupling between appearance and motion, and the motions in the generated videos do not sufficiently resemble those in the reference videos, with notably small motion amplitudes.

Tune-A-Video, which targets single-video customization and is based on the T2I model, suffers from poor inter-frame smoothness and severe appearance overfitting. Similarly, the few-shot motion customization method LAMP, also leveraging a T2I model, exhibits very poor temporal consistency and heavily relies on the quality of the initial frame. Compared to other methods, LAMP requires more reference videos and training iterations to achieve relatively better results. MotionDirector also encounters challenges with appearance overfitting, often generating unrealistic scenarios such as a panda on a skateboard dressed in human attire. Moreover, it exhibits insufficient modeling of motion patterns, resulting in videos with diminished motion magnitudes and deviations from the observed motion in reference videos.

Our method, however, demonstrates superior ability to accurately capture motion patterns in both one-shot and few-shot motion customization scenarios. 
Additionally, one-shot methods sometimes fail to discern whether to learn camera movement or foreground motion. 
In contrast, few-shot methods can leverage inductive biases derived from multiple videos, better capturing common motion patterns. This allows the temporal transformer to focus on foreground action rather than camera movements.

\begin{figure}[t]
\begin{center}
\includegraphics[width=1\linewidth]{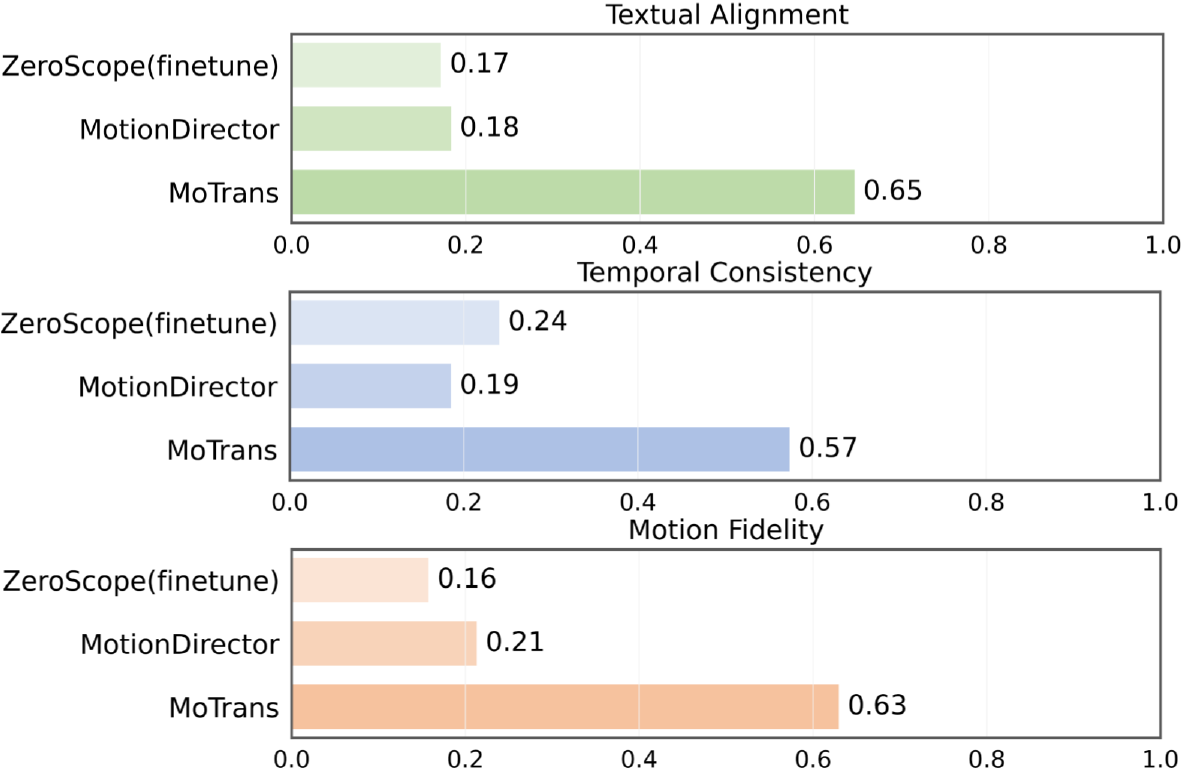}
\end{center}
   \caption{User study. For each metric, the percentages attributed to all methods sum to 1. MoTrans accounts for the largest proportion, indicating that the videos generated by our method exhibit superior text alignment, temporal consistency, and the closest resemblance to the reference video.}
\label{fig:user_study}
\end{figure}

\begin{figure}[t]
\begin{center}
\includegraphics[width=1\linewidth]{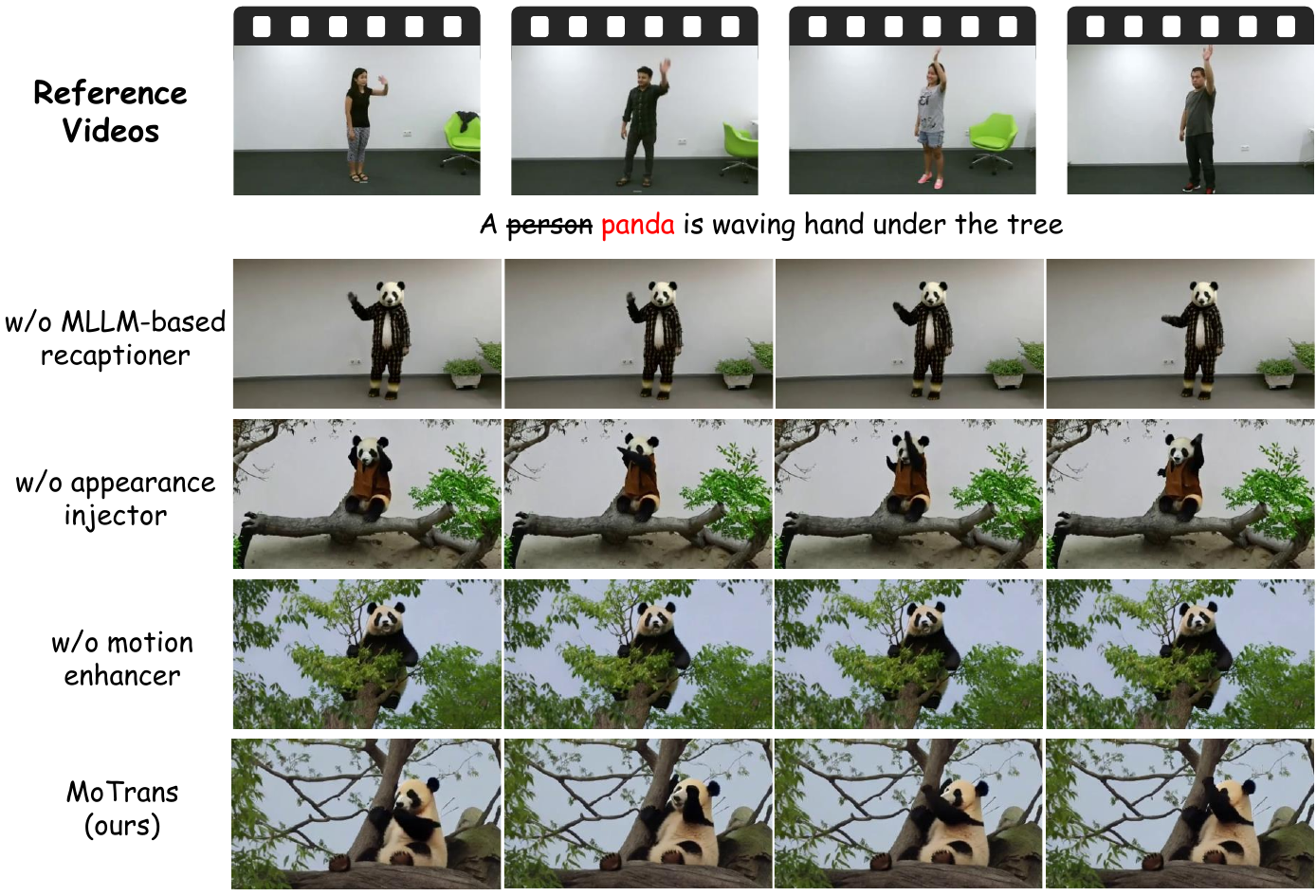}
\end{center}
   \caption{Qualitative results of the ablation study. Given several reference videos, Motrans can learn motion patterns from reference videos without appearance overfitting.}
\label{fig:ablation}
\end{figure}

\begin{figure*}[t]
\centering
\includegraphics[width=1\textwidth]{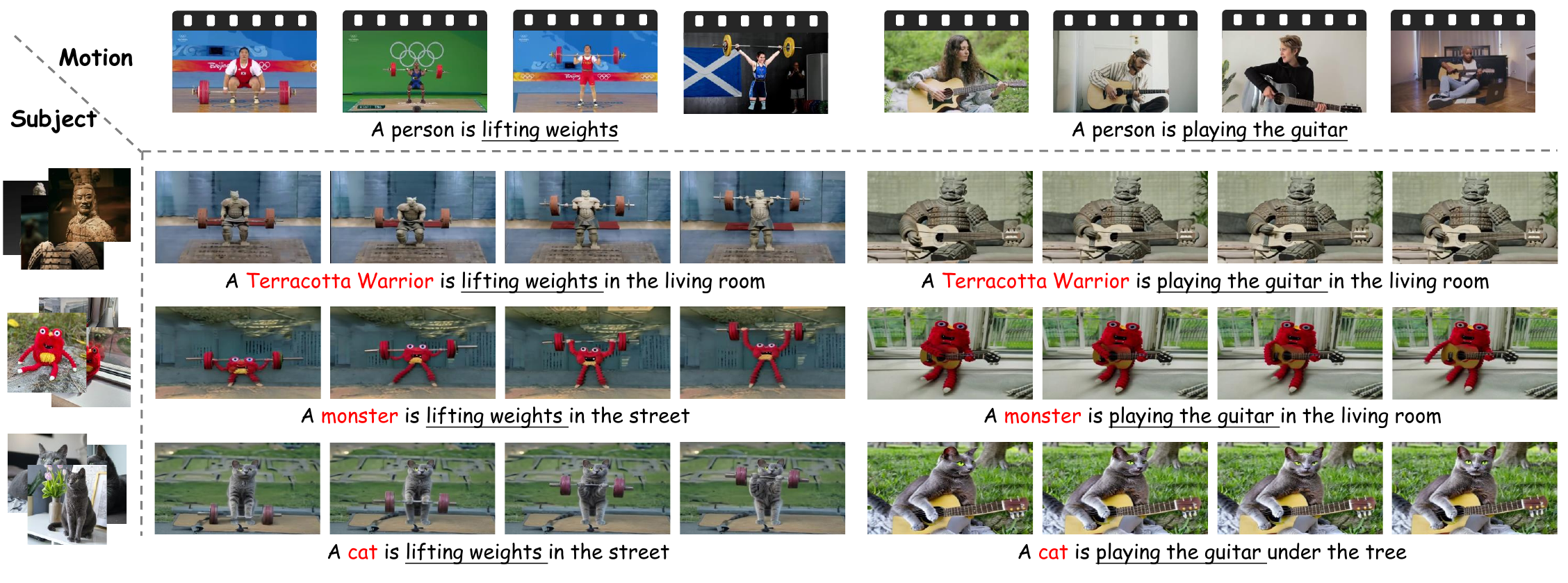}
\caption{Customized video generation with specific subjects and motions. The two-stage training strategy allows for the motion transfer~{(top)} from the reference video to the subject specified by exemplar images~{(left)}.}
\label{fig:subject_motion}
\end{figure*}

\noindent
\textbf{Quantitative Evaluation.}
As illustrated in Table~\ref{tab:quantitative}, when only a single reference video is provided, both Tune-a-Video and the fine-tuned Zeroscope exhibit higher motion fidelity but lower entity alignment. This is primarily attributed to the severe appearance overfitting, which leads to pronounced similarities in both appearance and motion to the reference video. Consequently, these methods fail to synthesize the new subject specified in the prompt, which is also demonstrated in Fig.~\ref{fig:qualitative}. In general, our method achieves a good balance between MoFid and CLIP scores, ensuring the generated results have the same motion as the reference videos and consistent subjects with the prompts.
When multiple reference videos are provided, our approach outperforms other methods across all evaluated metrics. Notably, it achieves high levels of text alignment and motion fidelity, showcasing our method's capability to effectively learn motion patterns from reference videos while avoiding overfitting the appearance information.

\noindent
\textbf{User study.}
Automatic metrics like CLIP-T have limitations in fully reflecting human preferences, hence, we conduct user studies to further validate our method. 
We collect 1536 sets of answers from 32 participants, with each completing a questionnaire containing 48 sets of questions. Participants are asked to pick the best video through answering the following questions: (1) which video better aligns with the textual description? (2) which video is smoother, and with fewer flickering? (3) which video's motion is more similar to that in the reference video without resembling its appearance?
Considering the significantly inferior performance of the T2I-based model LAMP, our comparison primarily focused on MoTrans versus the other methods. The results, shown in Fig.~\ref{fig:user_study}, reveal that our method consistently outperforms the others across all metrics, aligning more closely with human intuition.

\subsection{Ablation Study}
We conduct ablation studies to demonstrate the efficacy of the key modules introduced in this paper. Specifically, the MLLM-based recaptioner and the appearance injector leverage the prior knowledge of multimodal sources, i.e., textual and visual modalities, to address the challenge of coupling between appearance and motion. As illustrated in Table~\ref{tab:ablation} and Fig.~\ref{fig:ablation}~(rows 1 and 2), the absence of either the MLLM-based recaptioner or the appearance injector leads to a performance drop in both CLIP-T and CLIP-E, alongside high motion fidelity. This suggests a severe overfitting to both appearance and motion. Additionally, without the motion enhancer, the model struggles to synthesize the specific motion depicted in the reference videos, but with the introduction of the first two modules, it can synthesize the specified subject. In comparison, our method effectively mitigates appearance overfitting while ensuring motion fidelity as much as possible. Each module we propose significantly contributes to the improvement of the final generation results.

\subsection{Application}
\noindent
\textbf{Video customization with both subject and motion.}
Benefitting from the two-stage training strategy of our approach, appearance and motion can be learned separately through the spatial and temporal transformers within a UNet. As depicted in Fig.~\ref{fig:subject_motion}, we simultaneously customize the subject depicted in an image set and the motion specified by a video set. The customization results demonstrate that our method does not suffer from appearance overfitting to the training data and can successfully enable a specific animal or inanimate object to perform a human-centric motion.
\section{Conclusion}
We propose MoTrans, a customized motion transfer method that effectively transfers a specific motion pattern from reference videos to diverse subjects. By integrating multimodal appearance priors, encompassing both visual and textual modalities, our approach mitigates the issue of coupling between motion patterns in synthesized videos and the limited appearance contained in reference videos. Additionally, our method employs dedicated residual embeddings to accurately represent the specific motion pattern inherent in the reference videos.
Compared with existing methods, our method demonstrates superior capabilities in customizing motion and decoupling appearance, and it also supports the simultaneous customization of subjects and motions.

However, our method still encounters some limitations. First, our method focuses on transferring human-centric motion to objects with limbs. Some motions, like running, are unsuitable for objects without limbs, such as fish. 
Second, it is currently optimized for short video clips of 2-3 seconds and faces challenges in generating longer sequences. 
Our future work will aim to address these limitations and expand the applicability of our method to more complex and practical scenarios.
\begin{acks}
This work is partially supported by the National Natural Science Foundation of China, No. 62106036, No. U23B2010, No. 62293540, No. 62293542, and Dalian City Science and Technology Innovation Fund No. 2023JJ11CG001.
\end{acks}

\bibliographystyle{ACM-Reference-Format}
\bibliography{ref}

\clearpage
\appendix
\section{Details of Benchmarks}
To the best of our knowledge, there are currently no unified benchmarks for motion customization tasks.
Most representative motion customization methods~\cite{motiondirector,lamp,motioncrafter} typically select 8-12 different types of motions from UCF101~\cite{ucf101}, UCF Sports Action~\cite{ucfsports}, NTU RGB+D~\cite{ntuRGB+D120} for evaluation, covering a wide range of human-centric sports and daily activities. Following this setting, we have constructed our benchmark.
Considering that the existing datasets contain many simple motions with small movement amplitudes, such as walking frontally and snapping fingers, we do not directly use them for evaluation. Instead, we select videos with larger movement amplitudes and higher quality to enhance the diversity and complexity of our benchmark. We have carefully selected 12 motions including bowing, clapping, skateboarding, drinking water, lifting weights, kicking something, playing golf, riding a bicycle, pointing to something, playing the guitar, waving hand, and wiping face. 

\section{Analysis and Discussions}
\subsection{Discussion on Motion Fidelity Metric}
Considering the diverse composition of our benchmark, which includes not only sports actions but also various limb movements that accentuate dynamic processes, conventional metrics focused on single frame-to-text alignment, such as Textual Alignment~(CLIP-T) and Entity Alignment~(CLIP-E), may not provide sufficient measurement for motion quality. 
For instance, given the prompt "A tiger is drinking water in the forest", existing video foundation models often generate videos showing a tiger merely standing by a lake. While such outputs might achieve high CLIP-T and CLIP-E scores and ostensibly align with the textual description, they frequently misrepresent the specific motion in reference videos.
To address this challenge, we have introduced a novel metric named Motion Fidelity~(MoFid). This metric leverages the advanced video understanding capabilities of VideoMAE~\cite{videomae} to quantitatively assess how well the motion in generated videos matches the motion observed in the training dataset. 

Considering the different action types covered by VideoMAE and our method, this mismatch potentially leads to inaccuracies in motion type prediction. For instance, a video generated in response to the prompt "an alien is bowing" might be interpreted by VideoMAE as depicting "robot dancing", while a video produced for "a cat is wiping its face" could be erroneously categorized as "cat petting". Such discrepancies underscore the limitations of using straightforward classification accuracy to measure Motion Fidelity. Instead, we employ the cosine similarity of video representations to measure Motion Fidelity as mentioned in the main manuscript. 

\begin{figure}[t]
\begin{center}
\includegraphics[width=1\linewidth]{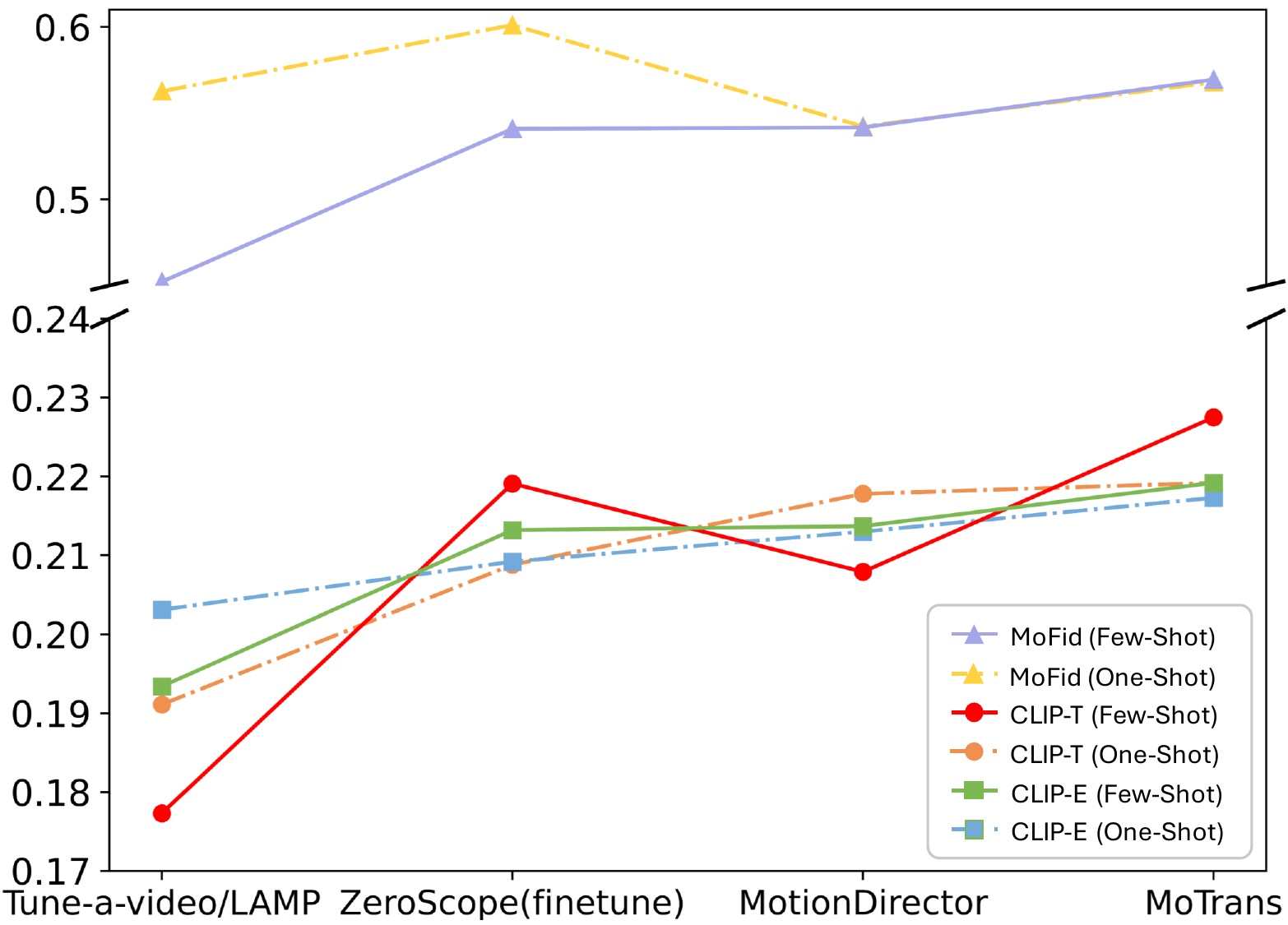}
\end{center}
   \caption{Relationship between Text/Entity Alignment and Motion Fidelity. A high MoFid score coupled with low CLIP-T and CLIP-E scores indicate that the synthesized video's appearance is excessively fitted to the reference video, resulting in a lack of appearance diversity.}
\label{fig:analysis_motion_fid}
\end{figure}

\begin{figure*}[ht]
\begin{center}
\includegraphics[width=1\linewidth]{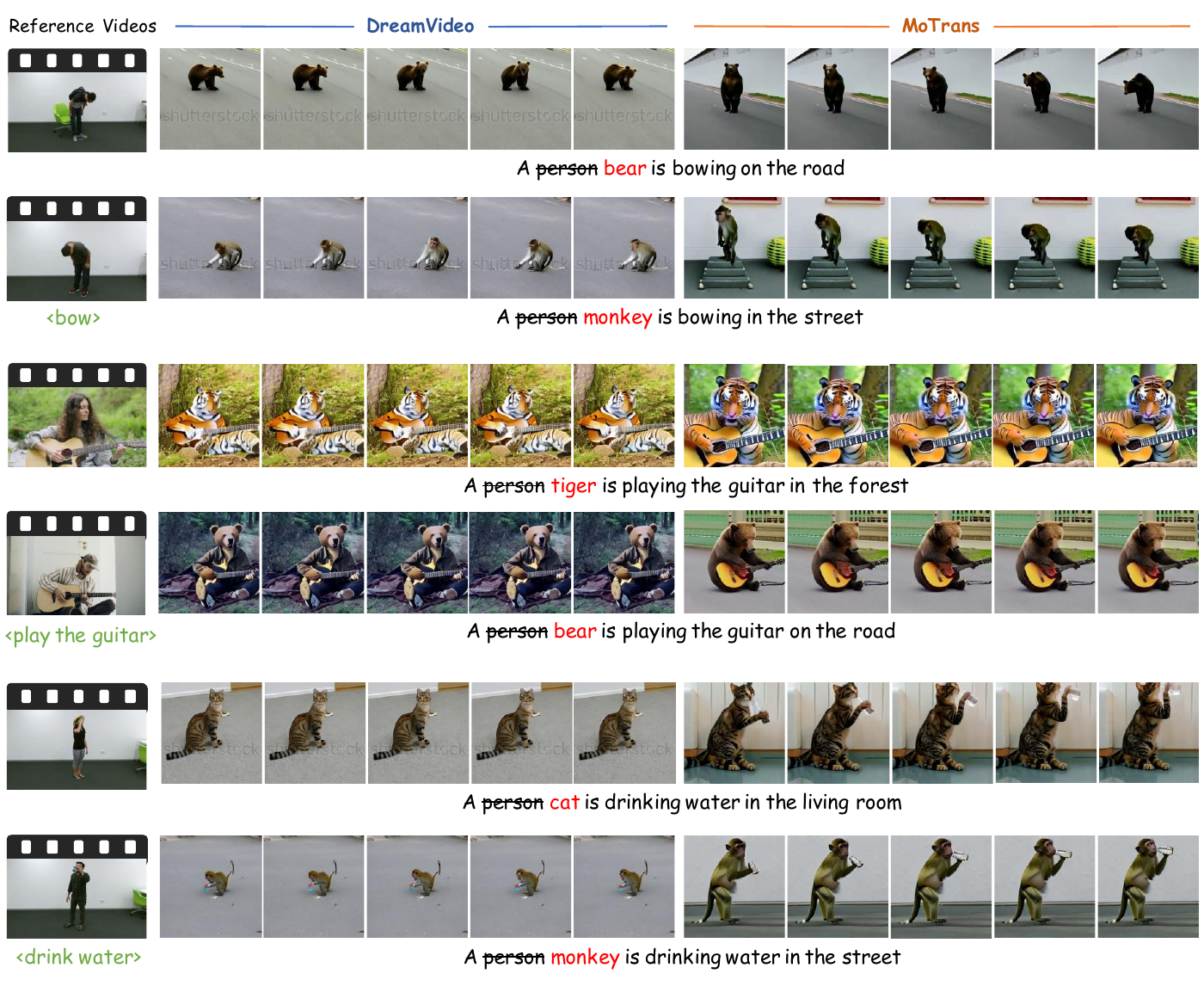}
\end{center}
   \caption{Qualitative comparisons between MoTrans and DreamVideo.}
\label{fig:motrans_vs_dreamvideo}
\end{figure*}

\subsection{Tradeoff between Text/Entity Alignment and Motion Fidelity}
We aim to explore the correlation between Text/Entity Alignment and Motion Fidelity, as depicted in Fig.~\ref{fig:analysis_motion_fid}. When a single reference image is provided, the finetuned ZeroScope achieves the highest Motion Fidelity score, yet its CLIP-T and CLIP-E metrics score the lowest. These results indicate a high similarity in both appearance and motion between the synthesized video and the reference video. In other words, while finetuned ZeroScope may accurately model the motion in the reference videos, it fails to generate the new context or entity implied by the prompts, thus limiting the overall creativity and diversity of the generated content.

Additionally, the videos synthesized by the few-shot methods align more closely with the text prompt, as evidenced by higher CLIP-E score. This suggests that the few-shot setting, compared to the one-shot setting, is better at synthesizing the subject specified in the prompt and avoids replicating appearances from the reference video.
Based on the analysis above, it can be concluded that although Motion Fidelity is a useful metric for assessing how consistently a video's motion matches that of the reference, it only provides a limited view of overall performance. High Motion Fidelity coupled with very low CLIP-T and CLIP-E scores typically indicates an overfitting to the reference's appearance. In contrast, our method consistently achieves high CLIP scores and Motion Fidelity in both one-shot and few-shot settings, demonstrating its effectiveness in modeling the motion pattern in reference videos without overfitting to their appearances. 

\begin{table}[t]
\renewcommand\arraystretch{1.1}
\centering
\caption{Quantitative comparison results of motion customization on multiple videos.
}
\label{tab:motrans_vs_dreamvideo}
\resizebox{0.48\textwidth}{!}{%
\begin{tabular}{ccccc}
\toprule
\textbf{Methods} & \textbf{CLIP-T}~($\uparrow$)  &  \textbf{CLIP-E}~($\uparrow$)  & \textbf{TempCons}~($\uparrow$)    & \textbf{MoFid}~($\uparrow$)  \\ 
\midrule
DreamVideo & 0.1791 & 0.2208  & 0.9680 & 0.4243 \\
MoTrans & \textbf{0.2168} & \textbf{0.2225} & \textbf{0.9776} & \textbf{0.5386} \\

\bottomrule
\end{tabular}
}
\end{table}

\section{Additional Results}
\subsection{Comparisons with DreamVideo}
We conduct additional qualitative and quantitative comparisons with DreamVideo~\cite{dreamvideo} to further demonstrate the superiority of our proposed MoTrans. DreamVideo employs an updated ModelScopeT2V model, which has been fine-tuned on its internal dataset at a resolution of 256. This fine-tuned version has not been made available to the public. Consequently, for our comparative analysis, we utilize the originally released ModelScopeT2V. DreamVideo is optimized for generating videos at a resolution of 256x256. For consistency and to enable a direct comparison, we also synthesize videos at this resolution. 
As shown in Table~\ref{tab:motrans_vs_dreamvideo}, DreamVideo can generate subjects specified by the prompt but often struggles to synthesize specific motions contained in the reference videos and the context specified by the prompt. Correspondingly, while its CLIP-E score is relatively high and comparable to our method, there is a significant difference in its CLIP-T and MoFid scores when compared to ours. The visual results in Fig.~\ref{fig:motrans_vs_dreamvideo} further confirm these observations. Compared to DreamVideo, our method demonstrates superior motion modeling capabilities.




\subsection{More Qualitative Results}
\begin{figure*}[h]
\centering
\includegraphics[width=1\textwidth]{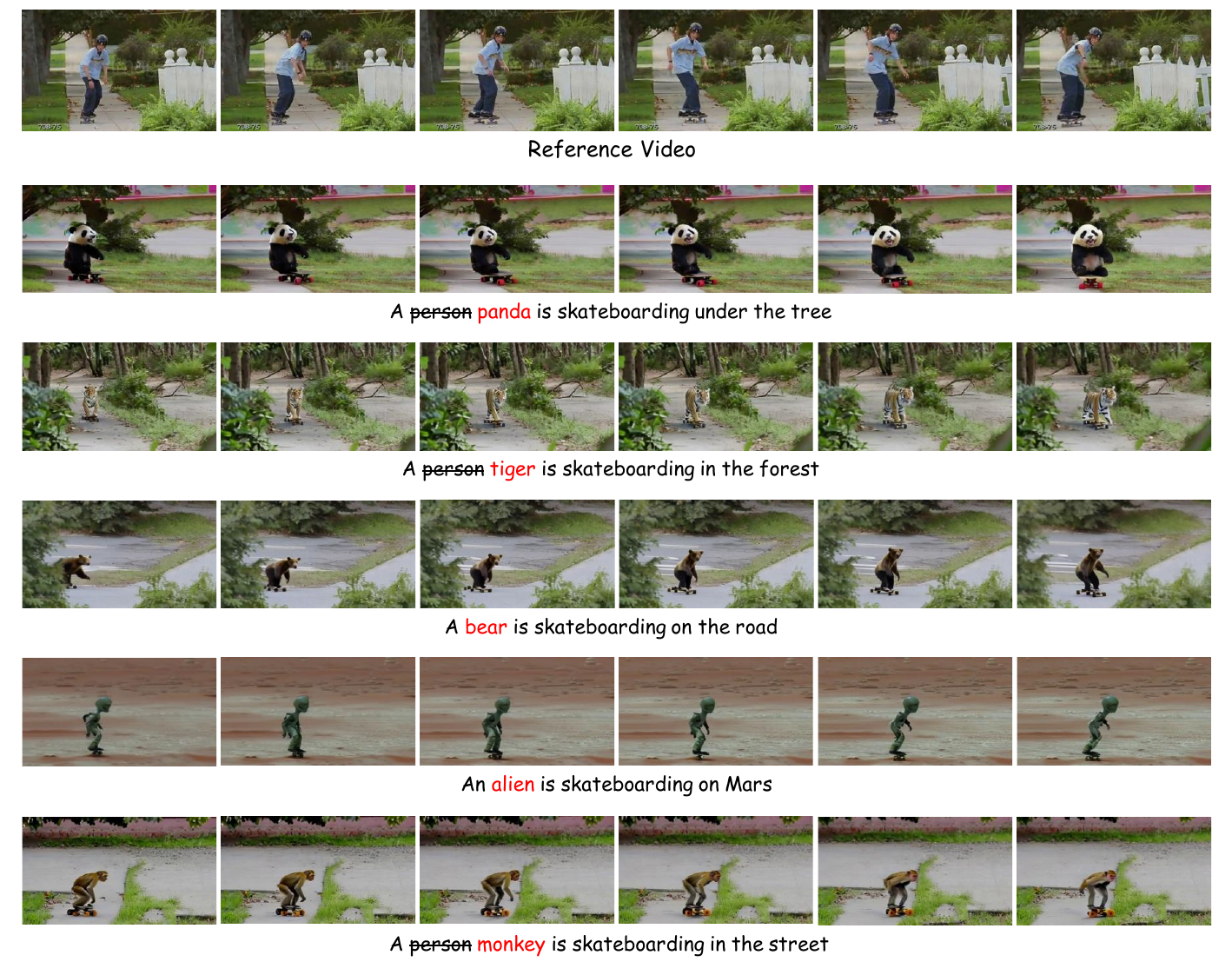}
\caption{Results of motion customization of the proposed MoTrans on single reference video. The first row specifies the reference video, showcasing a woman performing a skateboarding tic-tac action. Then the motion is transferred to various subjects specified by the new prompt.}
\label{fig:addition_one-shot}
\end{figure*}

\begin{figure*}[!t]
\centering
\includegraphics[width=1\textwidth]{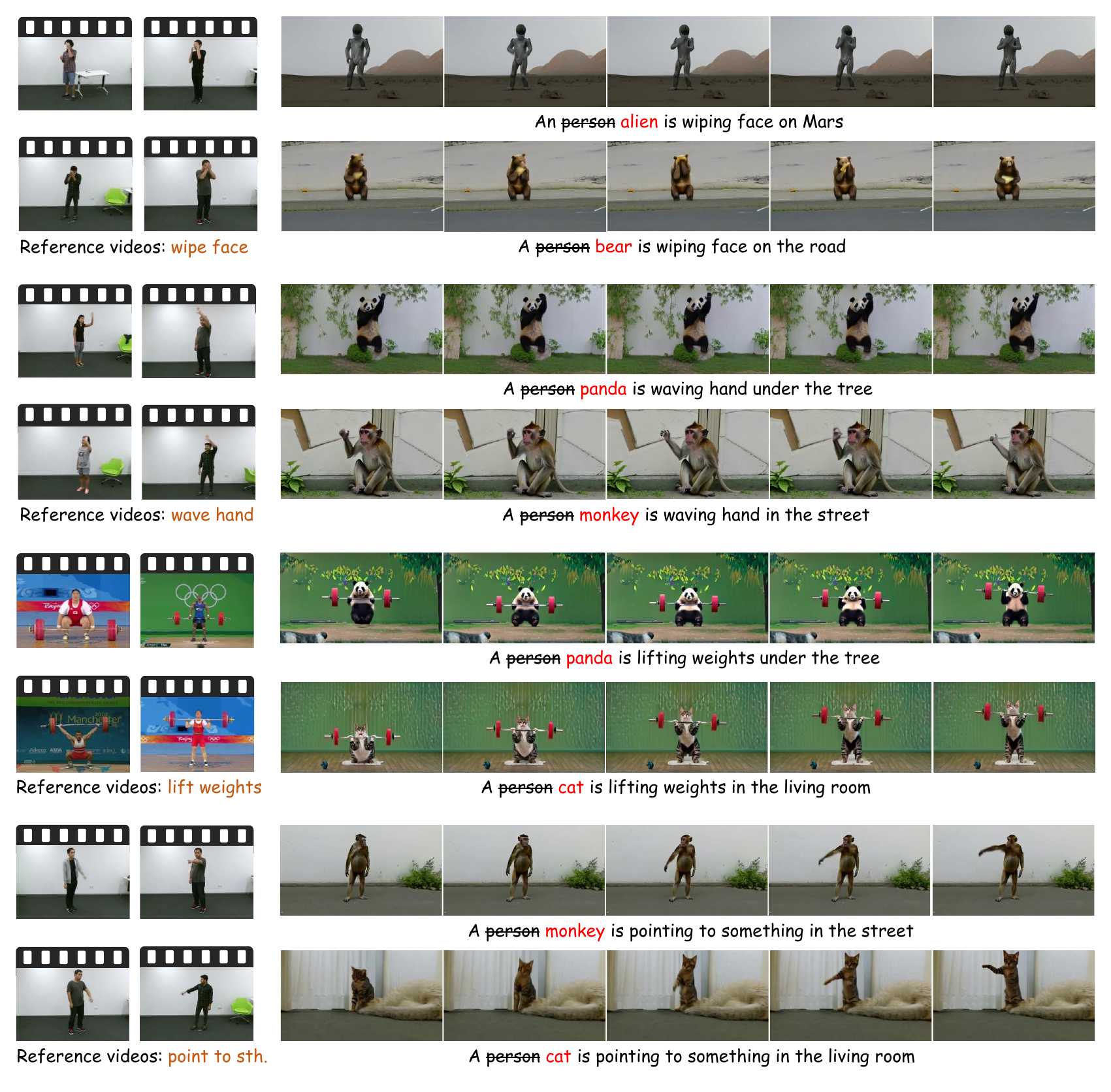}
\caption{Results of motion customization of the proposed MoTrans on multiple reference videos. The left side displays the reference video, while the right side shows the results of transferring the motion from the reference video to new subjects.}
\label{fig:addition_few_shot}
\end{figure*}

\begin{figure*}[!t]
\centering
\includegraphics[width=0.96\textwidth]{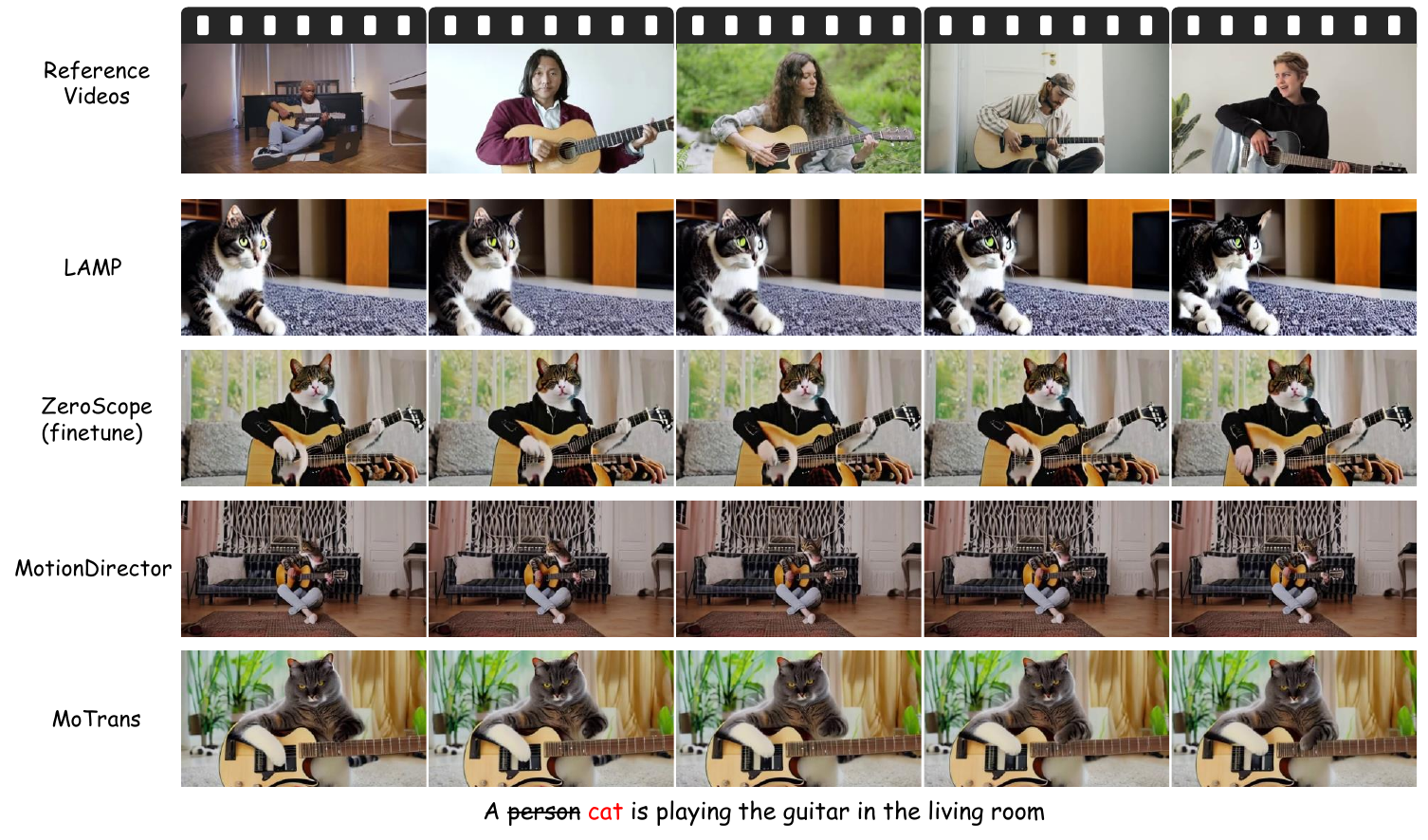}
\caption{Additional qualitative comparisons on customized motion transfer given multiple reference videos.}
\label{fig:addition_comparison}
\end{figure*}

\begin{figure*}[!t]
\centering
\includegraphics[width=0.96\textwidth]{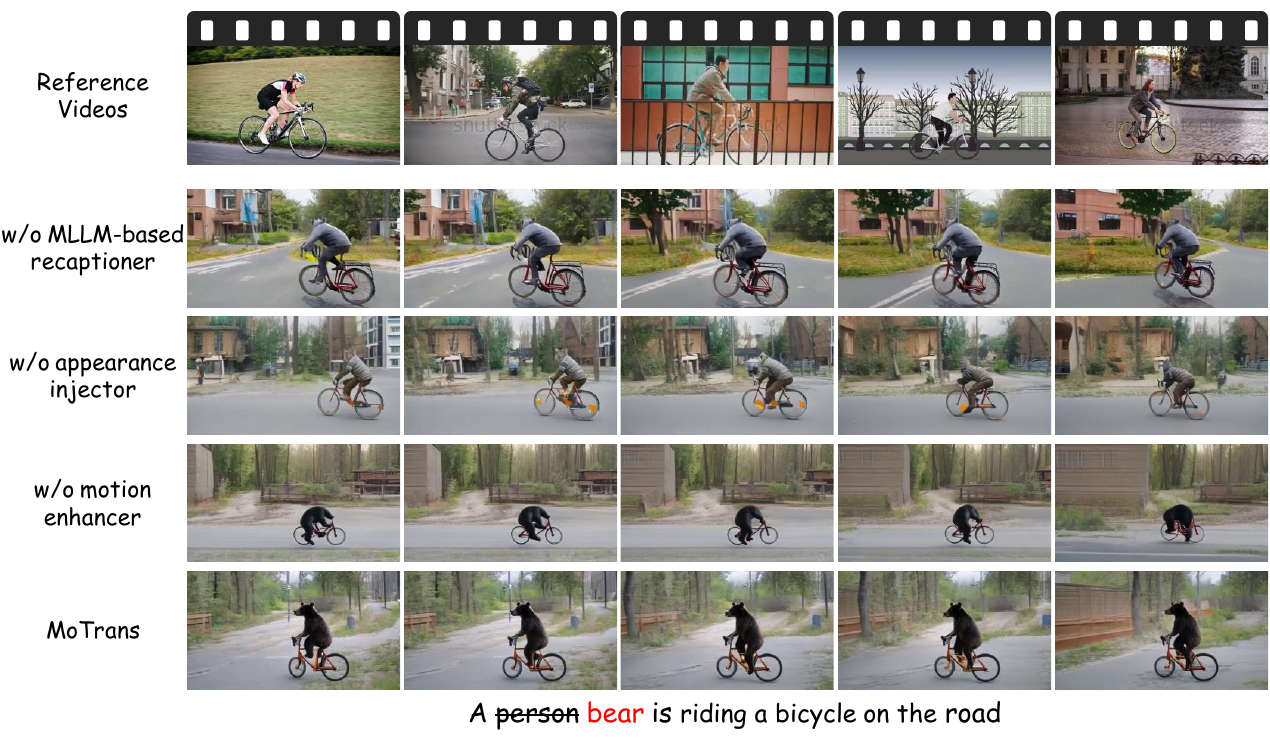}
\caption{Additional ablation study.}
\label{fig:addition_ablation}
\end{figure*}

Fig.~\ref{fig:addition_one-shot} and~\ref{fig:addition_few_shot} respectively present additional video examples synthesized by our approach, given a single reference video and multiple reference videos. We further compare our approach with other baselines, as demonstrated in Fig.~\ref{fig:addition_comparison}. Supplementary results from ablation studies are presented in Fig.~\ref{fig:addition_ablation}.

\end{document}